\documentclass{article}

\usepackage{arxiv}

\usepackage[utf8]{inputenc} 
\usepackage[T1]{fontenc}    
\usepackage{url}            
\usepackage{amsfonts}       
\usepackage{nicefrac}       
\usepackage{microtype}      
\usepackage{lipsum}
\usepackage{graphicx}

\usepackage{graphicx}
\usepackage{booktabs}
\usepackage{multirow}
\usepackage{mathtools}
\usepackage{pgfplots}
\usepackage{colortbl}
\usepackage{xcolor}
\usepackage{wrapfig}
\usepackage{makecell}

\usepgfplotslibrary{groupplots}
\usepackage{subcaption}

\usepackage{xcolor} 
\usepackage[normalem]{ulem}
\usepackage[accsupp]{axessibility}  
\usepackage{hyperref}
\usepackage{orcidlink}
\usepackage{cleveref}

\title{Optimizing Multi-Modal Models for Image-Based Shape Retrieval: The Role of Pre-Alignment and Hard Contrastive Learning}

\author{
Paul Julius Kühn\orcidlink{0000-0003-2458-0030} \\
Fraunhofer IGD, 64283 Darmstadt, Germany \And
Cedric Spengler\orcidlink{0009-0006-2323-5795} \\ 
Fraunhofer IGD, 64283 Darmstadt, Germany \And
Michael Weinmann\orcidlink{0000-0003-3634-0093} \\
Delft University of Technology, 2628 CD Delft, Netherlands
\And
Arjan Kuijper\orcidlink{ 0000-0002-6413-0061} \\
Fraunhofer IGD, 64283 Darmstadt, Germany \And
Saptarshi Neil Sinha\orcidlink{0000-0001-6637-0379} \\
Fraunhofer IGD, 64283 Darmstadt, Germany
}

\begin{document}
\maketitle
\begin{abstract}
  Image-based shape retrieval (IBSR) aims to retrieve 3D models from a database given a query image, hence addressing a classical task in computer vision, computer graphics and robotics. Recent approaches typically rely on bridging the domain gap between 2D images and 3D shapes based on the use multi-view renderings as well as task-specific metric learning to embed shapes and images into a common latent space. In contrast, we address IBSR through large-scale multi-modal pretraining and show that explicit view-based supervision is not required. Inspired by pre-aligned image–point-cloud encoders from ULIP and OpenShape that have been used for tasks such as 3D shape classification, we propose the use of pre-aligned image and shape encoders for zero-shot and standard IBSR by embedding images and point clouds into a shared representation space and performing retrieval via similarity search over compact single-embedding shape descriptors. This formulation allows skipping view synthesis and naturally enables zero-shot and cross-domain retrieval without retraining on the target database. We evaluate pre-aligned encoders in both zero-shot and supervised IBSR settings and additionally introduce a multi-modal hard contrastive loss (HCL) to further increase retrieval performance. Our evaluation demonstrates state-of-the-art performance, outperforming related methods on $Acc_{Top1}$ and $Acc_{Top10}$ for shape retrieval across multiple datasets, with best results observed for OpenShape combined with Point-BERT. Furthermore, training on our proposed multi-modal HCL yields dataset-dependent gains in standard instance retrieval tasks on shape-centric data, underscoring the value of pretraining and hard contrastive learning for 3D shape retrieval. The code will be made available via the project website.
 
\end{abstract}
\section{Introduction}
\begin{figure}[htb!]
    \centering
    \includegraphics[width=\linewidth]{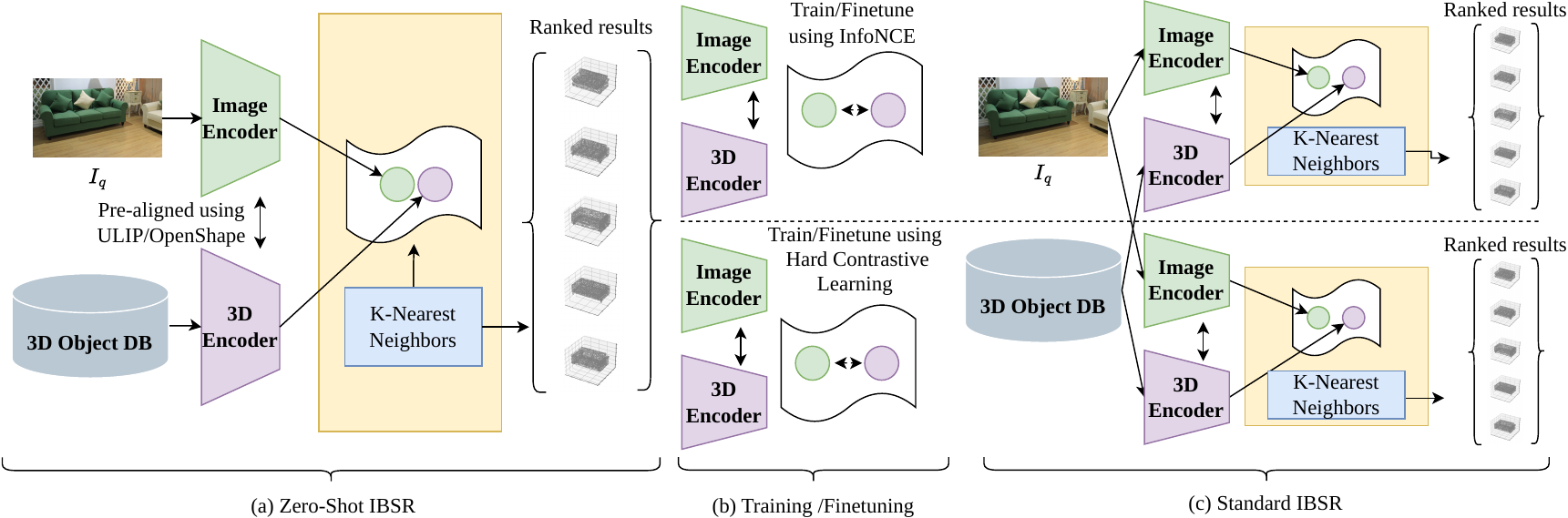}
    \caption{Training pipeline with multiple modalities. Zero-shot retrieval (a) uses pre-aligned image- and shape-encoders trained with multi-modal contrastive losses\cite{xue2023ulip}\cite{ulip2}\cite{liu2023openshape}. Training or fine-tuning (b) of image- and shape-encoders using either InfoNCE or hard contrastive learning enables standard retrieval (c) }.
    \label{fig:pipeline}
\end{figure}

Retrieving 3D shapes from visual observations is a fundamental problem in computer vision with applications in various sectors like e-commerce, inventory management, medical imaging, robotics, cultural heritage and many others. Given a query image, image-based shape retrieval (IBSR) aims to identify the corresponding 3D shape from a database-either at the instance level, recovering the exact depicted object, or at the class level, retrieving any shape of the same semantic category. IBSR requires bridging a fundamental domain gap between 2D images and 3D geometry. This entails two core components: extracting discriminative features from both modalities, and aligning these representations in a shared embedding space where semantically similar image-shape pairs lie close together. The dominant paradigm addresses this challenge by representing 3D shapes as collections of rendered 2D views~\cite{su2015mvcnn,grabner2018pose_estimation,fu2020hard_example,han2024ibsr}, enabling direct reuse of image encoders for both modalities. While effective, this approach discards native 3D geometric information and requires rendering shapes from multiple viewpoints at inference time that may not capture all relevant details depending on number and configuration of views. Recent advances in vision-language pre-training offer an alternative path. CLIP~\cite{radford2021clip} demonstrated that contrastive learning on large-scale image-text pairs yields aligned representations enabling zero-shot transfer to diverse downstream tasks. Subsequent work extends this paradigm to 3D: ULIP~\cite{xue2023ulip}, ULIP-2\cite{ulip2} and OpenShape~\cite{liu2023openshape} train point cloud encoders to align with CLIP's \cite{radford2021clip} pre-aligned embedding space that allow optimization of these two modalities using deep metric learning, achieving state-of-the-art performance on zero-shot 3D classification. By learning from image-text-point cloud triplets, these models produce geometry-preserving multi-modal representations with strong 3D classification performance. However, their effectiveness for IBSR—particularly zero-shot retrieval and retrieval-specific fine-tuning—remains unexplored. We propose a novel IBSR approach using pre-aligned multimodal encoders that operates directly on point clouds instead of multi-view renderings, eliminating view synthesis dependencies. Additionally, we introduce a multi-modal hard contrastive loss (HCL) that incorporates hard negative sampling for more effective differentiation between similar instances.
While hard negative sampling has proven effective in single-modality self-supervised learning \cite{robinson2021hcl,fu2020hard_example}, its application to IBSR is non-trivial due to the fundamental domain gap between 2D pixels and 3D points. Our novelty lies in adapting this objective to an asymmetric multi-modal setting, where the loss must simultaneously account for hard negatives in two distinct embedding distributions—image negatives for shape anchors and vice versa—to bridge the synthetic-to-real domain shift.
To the best of our knowledge, we are the first to apply pre-aligned multi-modal encoders and hard contrastive learning to IBSR, enabling robust cross-dataset retrieval with minimal retraining. We systematically evaluate pre-aligned encoders for IBSR under zero-shot and fine-tuned settings (Fig.~\ref{fig:pipeline}). We also adapt hard negative mining \cite{robinson2021hcl} to the asymmetric image--shape setting via a multi-modal hard contrastive loss (HCL). Rather than proposing a new architecture, we quantify when pre-alignment and hard negatives improve retrieval in practice. In light of these changes the main contributions are:
\begin{itemize}
    \item We employ and evaluate pre-aligned image and shape encoders for zero-shot and standard IBSR, extending pretraining/alignment techniques from classification to retrieval tasks and avoiding the need for multi-view rendering with its dependence on view selection and density.
    \item We propose a novel hard contrastive learning method for IBSR tasks, incorporating hard negative sampling to strengthen the model's discriminative capabilities, enabling better differentiation between similar instances.
    \item We perform quantitative analysis and ablation studies, 
    demonstrating that the proposed method outperforms prior methods across several benchmarks~\cite{wu2015modelnet, sun2018pix3d, yang2015compcars, krause2013stanfordcars} with with $Acc_{Top10}$ and $Acc_{Top1}$ performance approaching saturation (reaching $\approx \textbf{100\%}$). 
    Our proposed HCL improves standard retrieval, particularly for Point-BERT~\cite{yu2021pointbert} models.
    
\end{itemize}

\section{Related Work}
Connecting images to 3D shape repositories spans robotics, AR, and content creation. We review view-based, direct 3D, and multi-modal approaches.\\
\textbf{Multi-View Shape Representations.} The dominant paradigm represents 3D shapes through multiple rendered views. MVCNN~\cite{su2015mvcnn} pioneered this direction by aggregating CNN features across viewpoints. Follow-up work explores view selection strategies~\cite{lee2018view_sequence_learning}, sequential view modeling~\cite{he2019ngram}, and hard example mining~\cite{fu2020hard_example}. These methods typically employ metric learning to align rendered views with query images in a shared embedding space. Notably, by projecting shapes into the 2D domain prior to encoding, such approaches sidestep the challenge of learning directly from 3D geometry--limitation we address in this work.
Subsequent methods refine view aggregation through hierarchical grouping~\cite{feng2018gvcnn}, joint viewpoint-category learning~\cite{kanezaki2018rotationnet}, graph-based view relation modeling~\cite{wei2020viewgcn}, and differentiable viewpoint selection~\cite{hamdi2021mvtn}.
\\
\textbf{3D Shape Encoders.} An alternative to projecting shapes into 2D is to extract features directly from 3D representations. For point clouds, PointNet~\cite{qi2017pointnet} pioneered direct point cloud processing using shared MLPs and symmetric pooling, while PointNet++~\cite{qi2017pointnetpp} introduced hierarchical feature learning to capture local geometric structures. Subsequent works improved local feature aggregation through dynamic graph convolutions (DGCNN\cite{wang2019dgcnn}) or attention mechanisms (Point Transformer~\cite{zhao2021pointtransformer}, PCT~\cite{guo2021pct}). Recent studies show that even simple MLP architectures (PointMLP~\cite{ma2022pointmlp}) or revisited PointNet++ variants (PointNeXt~\cite{qian2022pointnext}) achieve competitive performance with modern training strategies. Self-supervised pretraining methods such as Point-BERT~\cite{yu2021pointbert} and Point-MAE~\cite{pang2022pointmae} further improve transferability through masked point modeling. For voxel-based representations, 3D ShapeNets~\cite{shapenet2015} and VoxNet~\cite{maturana2015voxnet} apply 3D convolutions on volumetric grids, while sparse convolutions~\cite{choy2019minkowski} enable efficient processing of high-resolution inputs. These approaches preserve geometric information but introduce a cross-modal alignment challenge: learned 3D embeddings must be aligned with image features to enable retrieval.\\
\textbf{Vision-Language Pre-training.} CLIP~\cite{radford2021clip} demonstrated that contrastive learning on large-scale image-text pairs yields powerful aligned representations enabling zero-shot transfer across diverse downstream tasks. CLIP employs modality-specific encoders: a Vision Transformer 
(ViT)~\cite{dosovitskiy2021vit} for images and a Transformer~\cite{vaswani2017transformer} 
for text. The model is trained with a symmetric cross-modal 
InfoNCE~\cite{oord2018infonce} objective that pulls matching image-text pairs 
together while pushing non-matching pairs apart. This paradigm has inspired extensions such as BLIP \cite{li2022blip}, SLIP \cite{mu2021slip}, and FILIP \cite{FILIP}, among others~\cite{ ScalingUp, SupervisionExistsEverywhere, Sigmoid4VLM, DemystifyingCLIP, sun2023evaclipimprovedtrainingtechniques, schuhmann2022laion}, as well as adaptations to additional modalities~\cite{ulip2, xue2023ulip, liu2023openshape, zhou2023uni3d}.
Beyond image-text retrieval, vision-language pre-training has enabled various cross-modal retrieval tasks. Composed Image Retrieval (CIR) \cite{zheng2018cbirsurvey} combines reference images with modification text to retrieve target images, leveraging similar contrastive objectives and pre-aligned embeddings. 
\\
\textbf{3D-Language Alignment.} Recent work extends vision-language alignment to 3D shapes. Early approaches such as PointCLIP~\cite{zhang2022pointclip} transfer CLIP's 2D knowledge to 3D by projecting point clouds into multi-view depth maps without 3D-specific training, with PointCLIP V2~\cite{zhu2023pointclipv2} improving upon this using realistic projections and GPT-generated prompts.
Rather than relying on projection-based transfer, ULIP~\cite{xue2023ulip, ulip2} trains point cloud encoders to project into CLIP's pre-aligned image-text embedding space using triplets of rendered images, text descriptions, and point clouds. By freezing the pre-trained CLIP \cite{radford2021clip} encoders and learning only the 3D branch, ULIP \cite{xue2023ulip} efficiently leverages existing vision-language alignment. OpenShape~\cite{liu2023openshape} scales this approach to larger datasets including Objaverse~\cite{objaverse2023}, while Uni3D~\cite{zhou2023uni3d} further explores unified 3D representations at scale. Point-Bind~\cite{guo2023pointbind} extends this paradigm to additional modalities including audio, and Cap3D~\cite{luo2023cap3d} contributes a scalable automatic 3D captioning pipeline for generating large-scale training data.
\noindent While these embeddings theoretically enable IBSR, prior work focused solely on zero-shot classification. Unlike multi-view methods \cite{su2015mvcnn, liu2019mvobjectretrieval, hamdi2021mvtn} requiring rendered 2D representations, we leverage pre-aligned point cloud encoders directly, preserving native 3D geometry. Furthermore, we investigate how hard contrastive learning \cite{robinson2021hcl}, which has proven effective in other metric learning contexts but has not been applied to multi-modal IBSR, can enhance instance-level discrimination in this setting. Current multi-modal 3D models typically rely on the symmetric InfoNCE loss~\cite{oord2018infonce}, which treats all negatives within a mini-batch as equally informative. We extend the theoretical framework of Robinson et al. \cite{robinson2021hcl}—originally designed for single-modality contrastive learning—to the multi-modal retrieval domain. Unlike previous works that focus on class-level alignment for classification, our HCL is specifically formulated to handle the cross-modal instance-level discrimination required for high-fidelity shape retrieval.
\section{Methodology}
\begin{figure}[htb!]
    \includegraphics[width=\textwidth]{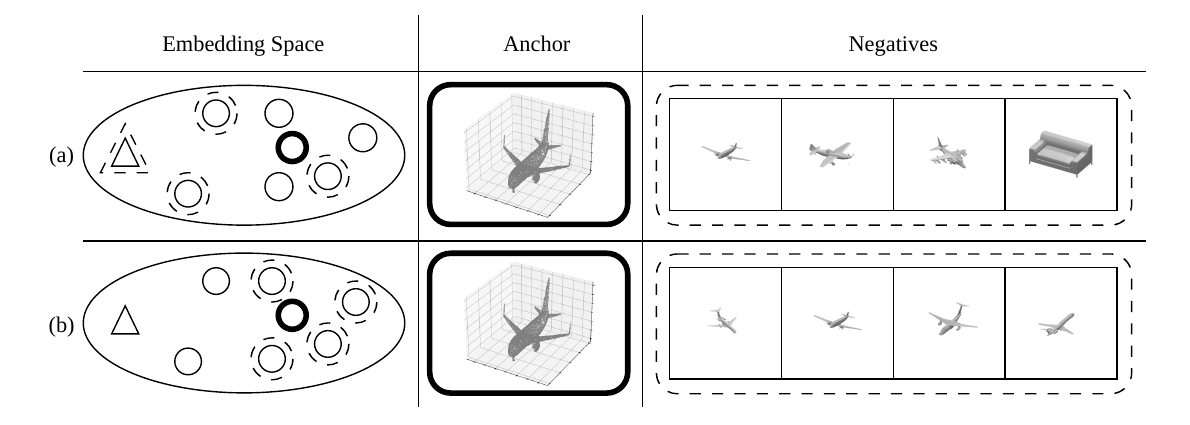}
    \caption[Motivation for Hard-Negative Learning]{
        Comparison of \textit{(a)} random sampling vs. \textit{(b)} hard negative sampling~\cite{robinson2021hcl}. Random sampling may yield out-of-class samples (e.g., couch $\bigtriangleup$ vs. airplane $\bigcirc$), producing overly easy negatives or even false negatives \textit{(a)}, while hard negative sampling ensures true hard negatives close to the anchor $\pmb{\bigcirc}$ \textit{(b)} \textit{(inspired by Fig. 1 of \cite{robinson2021hcl})}.}
    \label{fig:motivation_hard_negative_learning}
\end{figure}

Image-based shape retrieval (IBSR) retrieves 3D shapes from a query image, evaluated at instance or class level. Unlike multi-view methods~\cite{yang2021ibsr,fu2020hard_example,song2024scibsr}, we leverage pre-aligned encoders from ULIP~\cite{xue2023ulip,ulip2} and OpenShape~\cite{liu2023openshape}, enabling compact representations and zero-shot cross-domain retrieval without retraining. We further enhance performance through fine-tuning 
with hard-contrastive learning \cite{robinson2021hcl} shown in Fig. \ref{fig:motivation_hard_negative_learning} which illustrates how hard negative sampling yields near-anchor negatives that force finer discrimination and improve robustness, whereas random sampling sometimes produces easy or false negatives that degrade learning.
\subsection{IBSR Pipeline}
Our pipeline (Fig.~\ref{fig:pipeline}) consists of three components: an image 
encoder $f_I$, a point cloud encoder $f_P$, and a $k$-nearest neighbors ($k$-nn) 
matcher. Both encoders map inputs to a shared embedding space where corresponding 
image-shape pairs are closely positioned. Given a shape database $D_S$, we embed 
all point clouds via $f_P$, build a $k$-nn index~\cite{k_nearest}, then retrieve 
the $k$ nearest shapes for a query image embedded by $f_I$. When $f_P$ has not 
seen $D_S$ during training, this constitutes zero-shot retrieval.
Following ULIP \cite{xue2023ulip, ulip2} and OpenShape \cite{liu2023openshape}, we use frozen pre-aligned image-text encoders 
(e.g., OpenCLIP~\cite{ilharco2021openclip}) and align only the point cloud 
encoder, simplifying training and avoiding forgetting.

\subsection{Design Choices}
Our approach differs from prior work~\cite{yang2021ibsr,fu2020hard_example,song2024scibsr} 
in two key aspects:

\noindent\textbf{Point cloud representation.} Unlike multi-view methods \cite{su2015mvcnn, lee2018view_sequence_learning, he2019ngram}, point 
clouds preserve explicit geometric information without requiring fusion of 
multiple feature tensors. This reduces data complexity (e.g., 60k values vs.\ 
1.5M for 10-view representations) while improving robustness to rotation and 
partial observations. Furthermore, avoiding the need for view synthesis avoids the dependence on the number and configurations of considered views.\\
\noindent\textbf{Pre-aligned encoders.} We consider pre-alignment as a specific form of pre-training whose objective is to embed heterogeneous modalities into a shared latent space while enforcing cross-modal alignment. By leveraging encoders aligned on 
large-scale data (e.g., LAION-5B~\cite{schuhmann2022laion}), we improve data 
efficiency and enable zero-shot and cross-domain retrieval without retraining.

\subsection{Image-based shape retrieval training}
\label{sec:training}
Pre-aligned encoders enable zero-shot retrieval on unseen shape datasets. 
To improve performance on domain-specific data, we tune the point cloud 
encoder while keeping the image encoder frozen—referred to as \emph{standard 
retrieval}. Since our task involves only shapes and images, we exclude the 
text modality.\\
\noindent\textbf{Tuple Creation}
fine-tuning requires point clouds paired with $M$ rendered images per shape, 
yielding $N$ point clouds with $M$ images each ($N$:$M$).\\
\noindent\textbf{Point cloud generation.}
We uniformly sample $P$ points ($P{=}8{,}000$ for ULIP~\cite{xue2023ulip}, $P{=}10{,}000$ for ULIP2~\cite{ulip2} and OpenShape~\cite{liu2023openshape}) from each 3D surface. RGB values are included when available $(P \times 6)$, as color improves performance~\cite{liu2023openshape}; otherwise, uniform color is applied. Point clouds are normalized and axis-aligned with the pre-trained encoder. We apply the following augmentations: random dropout, scaling, point shift, perturbation, and rotation.\\
\noindent\textbf{Image generation.}
We consider RGB images with the dimensions $H \times W \times C$. When not available, we render 12 views per shape ($30^{\circ}$ elevation, $30^{\circ}$ azimuth intervals) following~\cite{liu2023openshape,xue2023ulip}. Images are preprocessed (resizing, center cropping, normalization) per downstream encoder, using OpenCLIP~\cite{ilharco2021openclip} by default. We do not apply further augmentation. For further details, we refer to the supplementary material. These images serve only for tuple creation, not shape description.\\
\noindent\textbf{Training framework}
We finetune only the point cloud encoder $f_P$ while keeping $f_I$ frozen. Training follows the 
same contrastive objective as pre-training. Mini-batches of $N$ shape-image 
tuples $\{(p,im)\}^{1, \ldots, N}$ are constructed by augmenting each point cloud and 
randomly selecting one corresponding image. Embeddings from $f_P$ and $f_I$ 
are $\ell_2$-normalized and aligned via the multi-modal InfoNCE 
loss~\cite{oord2018infonce}:

\begin{multline}
    \label{eq:finetune_loss}
    \mathcal{L}_{P\rightarrow I}(\{(p,im)^N\}) = \frac{1}{N}\sum_{i=1}^N-\frac{1}{2}\log \frac{\exp(f_P(p_i)^\top f_I(im_i)/\tau)}{\sum^N_{j=1} \exp(f_P(p_i)^\top f_I(i_j)/\tau)}\\
    -\frac{1}{2}\log\frac{\exp(f_P(p_i)^\top f_I(im_i)/\tau)}{\sum^N_{j=1} \exp(f_P(p_j)^\top f_I(im_i)/\tau)}
\end{multline}
The temperature $\tau$ from Eq. \ref{eq:finetune_loss} is re-initialized rather than inherited from 
pre-training. Since $f_I$ is frozen, image embeddings are computed offline, 
significantly accelerating training.

\noindent\textbf{Hard negative learning} Standard contrastive learning samples negatives uniformly, risking uninformative gradients from easy negatives far from the anchor. We integrate hard negative 
sampling~\cite{robinson2021hcl} with the multi-modal version of InfoNCE introduced by CLIP \cite{radford2021clip}, using a 
parameterized distribution $q$ that up-weights negatives close to positive anchors 
(Fig.~\ref{fig:motivation_hard_negative_learning}). This improves instance-level 
discrimination without additional computational cost, as reweighting occurs 
within existing mini-batches.
\noindent Following Robsinson et. al. \cite{robinson2021hcl}, we modify the InfoNCE loss over a batch of positive pairs $\{x,x^+\}$ to train a parameterized encoder $f_\theta$ by replacing in-batch negatives with negatives sampled from distribution $q$ denoting the distribution of true negatives. Replacing uniform sampling with $Q$ negatives drawn from $q$, the objective  becomes:


\begin{equation}
    \label{eq:hcl_nce}
	\mathcal{L}(\{x,x^+\}^N) = -\frac{1}{N}\sum^N_{i=1} \log\frac{\exp f_\theta(x_i)^\top f_\theta(x_i^+)}{\exp f_\theta(x_i)^\top f_\theta(x_i^+) + Q \mathbb{E}_{x^-\sim q_\beta}[\exp f_\theta(x_i)^\top f_\theta(x^-)]}
\end{equation}




The novelty of our HCL (Eq.\ref{eq:mm_hcl}) is the symmetric cross-modal extension. In the IBSR context, a 'hard negative' is not just a similar image, but a 3D shape whose geometric embedding is deceptively close to the query image's visual embedding. Since the context of instance level retrieval omits the necessity to take care about false-negatives In Eq. \ref{eq:hcl_nce}, all samples from $q$ are true-negatives: $q_\beta^-(x^-) = q_\beta(x^-)$ (Eq.~\ref{eq:beta_sampling}). Extending to the multi-modal setting yields our hard contrastive loss (HCL) objective:
\begin{align}
    \label{eq:mm_hcl}
    & \mathcal{L}^{HCL}_{P\rightarrow I}(\{(p,im)^N\}) = \frac{1}{N}\sum_{i=1}^N \Bigg( \\ \nonumber
    & - \frac{1}{2} \log \frac{\exp(f_P(p_i)^\top f_I(im_i)/\tau)}{\exp(f_P(p_i)^\top f_I(im_i)/\tau) + Q\mathbb{E}_{im^-\sim q^{im}_\beta}[\exp(f_P(p_i)^\top f_I(im^-) / \tau)]} \\ \nonumber
    & - \frac{1}{2} \log \frac{\exp(f_P(p_i)^\top f_I(im_i)/\tau)}{\exp(f_P(p_i)^\top f_I(im_i)/\tau) + Q\mathbb{E}_{p^-\sim q^p_\beta}[\exp(f_P(p^-)^\top f_I(im_i)/\tau)]} \Bigg) \nonumber \\ \nonumber
\end{align}
where $\tau$ is the learnable temperature and $Q = N - 1$ is the number of negatives per batch of size $N$.



\begin{figure}[htb!]
    \centering
    \includegraphics[width=0.5\linewidth]{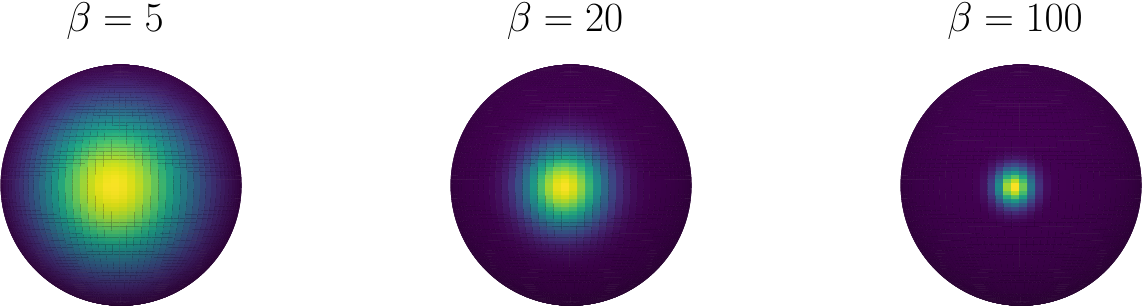}
    \caption{Von Mises-Fisher distribution on a unit hypersphere for varying 
    $\beta$. Higher $\beta$ increases concentration of $q_\beta$ and negative 
    hardness.}
    \label{fig:misis_fisher}
\end{figure}

\noindent\textbf{Modeling the negative distribution.}
By modeling the negative distributions $q_i^\beta$ (images) and $q_p^\beta$ (point clouds) separately (see Eq. \ref{eq:beta_sampling}), we force the model to resolve fine-grained ambiguities that are often ignored by standard multi-modal InfoNCE~\cite{radford2021clip}.
The distribution $q_\beta$ must lie on the latent hypersphere, concentrate 
around the anchor, and allow smoothness control via $\beta$. Following 
Robinson~et~al.~\cite{robinson2021hcl}, we model $q_\beta$ as an unnormalized 
von Mises-Fisher distribution using the anchor as mean direction and $\beta$ 
as concentration:
\begin{equation}
    \label{eq:beta_sampling}
    q_\beta(x^-) \propto \exp(\beta f(x)^\top f(x^-)) \cdot p(x^-)
\end{equation}
Fig.~\ref{fig:misis_fisher} illustrates how $\beta$ controls the concentration 
of negatives around the anchor. Further details are provided in the supplementary material.

\section{Evaluation}
In order to demonstrate the potential of our approach, we compare with different baseline approaches. 
\subsection{Baseline techniques used for comparison}
We compare against state-of-the-art methods, including ULIP~\cite{xue2023ulip}, ULIP2~\cite{ulip2}, and OpenShape~\cite{liu2023openshape}, using compatible encoders and datasets. Our results demonstrate that hard contrastive learning yields comparable or improved zero-shot retrieval with pre-aligned models, while standard retrieval with Point-BERT~\cite{yu2021pointbert} shows notable accuracy gains from our hard contrastive learning objective training (Tables~\ref{tab:zeroshot_ibsr},\ref{tab:standard_performance},\ref{tab:hcl_zeroshot}).
\subsection{Metrics}
We report Top-$k$ accuracy ($Acc_{\mathrm{Top}k}$) and mean average precision 
($mAP@10$). $Acc_{\mathrm{Top}1}$ measures whether the correct shape is 
retrieved at rank 1, while $Acc_{\mathrm{Top}10}$ captures whether it 
appears within the top 10 results. These metrics indicate retrieval 
success but not ranking quality. To address this, $mAP@10$ evaluates 
precision across ranks, providing a complementary assessment of 
retrieval effectiveness. CompCars \cite{yang2015compcars} and StanfordCars~\cite{krause2013stanfordcars} are single class datasets, hence we report only instance level results. For all other datasets, we provide both instance-level and class-level scores. Standard retrieval results for the Pix3D~\cite{sun2018pix3d} dataset do not present class-level scores in some cases (Tab. \ref{tab:standard_performance}), because baseline frameworks do not evaluate them; as a result, we cannot provide retrieval-gain comparisons.
\subsection{Datasets}
We evaluate on two dataset categories: \textit{shape-centered} collections 
with train/test splits by shape, and \textit{image-centered} IBSR benchmarks 
with splits by image (allowing shape overlap between splits). For shape-centered 
evaluation, we use ModelNet40~\cite{wu2015modelnet} and the LVIS~\cite{Gupta2019LVISAD}-annotated subset of Objaverse~\cite{objaverse2023} dataset following ULIP2~\cite{ulip2} and OpenShape~\cite{liu2023openshape} protocols. For IBSR, we evaluate on Pix3D~\cite{sun2018pix3d}, CompCars~\cite{yang2015compcars}, 
and StanfordCars~\cite{krause2013stanfordcars}. Table~\ref{tab:datasets} summarizes dataset statistics.
We evaluate two Point-BERT~\cite{yu2021pointbert} variants: Point-BERT(S)(5.1M params) and Point-BERT(L) (72.1M params).
\begin{wraptable}{r}{0.5\textwidth}
    \vspace{1pt}
    \centering
    \small
    \scalebox{0.7}{
    \begin{tabular}{llccc}
        \toprule
        Type & Dataset & Shapes & Train Img & Test Img \\
        \midrule
        \multirow{3}{*}{Shape-centered}
          & ModelNet40~\cite{wu2015modelnet} & 12,311 & -- & -- \\
          & Objaverse-LVIS~\cite{objaverse2023} & 46k & -- & -- \\
          
        \midrule
        \multirow{3}{*}{Image-centered}
          & Pix3D~\cite{sun2018pix3d} & 322 & 2,648 & 2,470 \\
          & CompCars~\cite{yang2015compcars} & 94 & 3,798 & 1,898 \\
          & StanfordCars~\cite{krause2013stanfordcars} & 134 & 8,144 & 8,041 \\
          \bottomrule
    \end{tabular}}
    \caption{Dataset statistics.}
    \label{tab:datasets}
    \vspace{-55pt}
\end{wraptable}

\subsection{Implementation details}
All experiments were conducted on 2 NVIDIA A100 GPUs using AdamW. 
Learning rates are 
$5 \times 10^{-4}$ for Point-BERT~\cite{yu2021pointbert} and 
$1 \times 10^{-3}$ for SparseConv~\cite{choy2019minkowski} encoders.\\
\noindent\textbf{Standard retrieval.} Models are finetuned using the train split with hard contrastive or InfoNCE loss.
We train for 100 epochs on ModelNet40~\cite{wu2015modelnet} and 1000 epochs 
on other datasets. \textbf{Zero-shot retrieval.} Models are trained for 100 epochs on 
ShapeNet~\cite{shapenet2015} with batch size 200, following 
OpenShape~\cite{liu2023openshape}. For the HCL, we use 
initial $\beta = 0.5$ with five-stage annealing~\cite{robinson2021hcl}. 

\subsection{Training modalities}
We employ two different training modalities: Standard and zero-shot retrieval.
\emph{Zero-shot retrieval} is defined as the ability of our pre-aligned image and point-cloud encoders—aligned via large-scale image–shape pairs following PointCLIP \cite{zhang2022pointclip} and ULIP \cite{xue2023ulip}—to retrieve 3D shapes for novel image queries, including entirely unseen shape datasets, without any further training. To evaluate zero-shot performance, we apply these fixed encoders to the retrieval task.
For \emph{standard retrieval}, we freeze the image encoder and fine-tune only the shape encoder on domain-specific shapes and rendered images to adapt to the target domain. The point-cloud encoder alignment itself proceeds in two stages: (1) a pre-alignment (pre-training) phase that generally brings the point-cloud encoder into correspondence with the image encoder, and (2) a domain-specific fine-tuning phase that specializes it on the target data. Hard-negative learning (Eq.~\ref{eq:mm_hcl}) is applied during both stages analyzing its effect on both \emph{zero-shot} and \emph{standard} retrieval modalities.
\subsection{Zero-Shot Retrieval with InfoNCE Baseline} \label{ssec:nce_zeroshot}
We evaluate zero-shot retrieval on ModelNet40~\cite{wu2015modelnet} and 
Objaverse-LVIS~\cite{objaverse2023}. As shown in Table~\ref{tab:zeroshot_sc}, 
OpenShape~\cite{liu2023openshape} models consistently outperform 
ULIP~\cite{xue2023ulip,ulip2} for IBSR. Performance improves with 
(1)~large-scale pre-training (Ensembled vs.\ ShapeNet~\cite{shapenet2015}-only) and 
(2)~increased model capacity. Point-BERT(L)~\cite{yu2021pointbert} 
trained via OpenShape ~\cite{liu2023openshape} achieves the best results.
Notably, all models perform significantly better at class-level than 
instance-level retrieval, except SparseConv~\cite{liu2023openshape}, 
which shows comparable discrimination at both levels despite lower 
overall accuracy.
For Objaverse-LVIS evaluation, models trained on the full Ensembled 
dataset are excluded due to data overlap. Point-BERT(S)~\cite{yu2021pointbert} trained on 
(excluding the LVIS~\cite{Gupta2019LVISAD} subset) achieves the best results, outperforming 
its ModelNet40 \cite{wu2015modelnet} performance by ${\sim}16\%$—likely due to exposure to 
similar Objaverse~\cite{objaverse2023} data during pre-training.
\begin{table}[htb!]
\centering
\resizebox{0.8\textwidth}{!}{%
\begin{tabular}{l|lll|ccc}
Dataset                        & Framework & Model        & Pre-train Data                                               & $Acc_{Top1}$       & $Acc_{Top10}$      & $mAP@10$           \\ \hline
\multirow{8}{*}{Modelnet40~\cite{wu2015modelnet}}    & ULIP~\cite{xue2023ulip}     & Point-BERT~\cite{yu2021pointbert}    & Shapenet~\cite{shapenet2015}                                                     & 0.3/9.5            & 2.1/35.9           & 0.7/15.6           \\
                               &           &              & Ensembled~\cite{liu2023openshape}                                                   & 1.6/22.8           & 6.4/61.1           & 2.9/30.3           \\
                               & OpenShape~\cite{liu2023openshape} & Point-BERT(S)~\cite{yu2021pointbert} & Shapenet~\cite{shapenet2015}                                                     & 18.7/71.1          & 55.2/90.7          & 29.1/73.9          \\
                               &           & Point-BERT(L)~\cite{yu2021pointbert} & Ensembled~\cite{liu2023openshape}                                                    & 	 \cellcolor{green!20}\textbf{30.5/97.2} & \cellcolor{green!20}\textbf{67.1/92.9} & \cellcolor{green!20}\textbf{41.6/80.1} \\
                               &           & SparseConv~\cite{choy2019minkowski}  & Shapenet~\cite{shapenet2015}                                                    & 9.5/63.8           & 38.0/88.2          & 17.1/67.4          \\
                               &           &              & Ensembled~\cite{liu2023openshape}                                                    & 17.9/18.0          & 51.0/52.6          & 27.6/27.7          \\
                               & ULIP2~\cite{ulip2}     & Point-BERT~\cite{yu2021pointbert}   & Ensembled~\cite{liu2023openshape}                                                    & 3.0/3.1            & 9.0/10.2           & 4.6/4.9            \\
                               &           & PointNeXt~\cite{qian2022pointnext}    & Ensembled~\cite{liu2023openshape}                                                    & 3.5/3.6            & 14.6/16.0          & 6.2/6.6            \\ \hline
\multirow{5}{*}{ObjaverseLVIS~\cite{objaverse2023}} & ULIP~\cite{xue2023ulip}      & Point-BERT~\cite{yu2021pointbert}    & Shapenet~\cite{shapenet2015}                                                     & 0.1/2.1            & 0.4/10.6           & 0.2/4.1            \\
                               & OpenShape~\cite{liu2023openshape} & Point-BERT(S)~\cite{yu2021pointbert} & Shapenet~\cite{shapenet2015}                                                     & 8.2/20.7           & 26.2/49.3          & 13.2/26.3          \\
                               &           & Point-BERT(L)~\cite{yu2021pointbert} & \begin{tabular}[c]{@{}l@{}}Ensembled~\cite{liu2023openshape}\\ (NoLVIS)\end{tabular} & \cellcolor{green!20}\textbf{46.9/69.4} & \cellcolor{green!20}\textbf{82.7/93.5} & \cellcolor{green!20}\textbf{58.4/69.6} \\
                               &           & SparseConv~\cite{choy2019minkowski}   & Shapenet~\cite{shapenet2015}                                                     & 2.6/12.6           & 11.6/36.5          & 4.9/17.5           \\
                               &           &              & \begin{tabular}[c]{@{}l@{}}Ensembled~\cite{liu2023openshape}\\ (NoLVIS)\end{tabular} & 24.7/49.6          & 60.6/84.4          & 34.9/55.0          \\ \hline
\end{tabular}%
}
\caption[\textit{Zero-Shot} results  for shape datasets]{Zero-shot retrieval results for shape-centered datasets (aligned using InfoNCE loss~\cite{oord2018infonce}) on ModelNet40~\cite{wu2015modelnet}, ObjaverseLVIS~\cite{objaverse2023}. \textit{(Notation: $\text{instance}/\text{class}$)}}
\label{tab:zeroshot_sc}
\end{table}
\begin{table}[htb!]
\centering
\resizebox{0.8\textwidth}{!}{%
\begin{tabular}{l|lll|ccc}
Dataset                       & Framework & Model        & Pre-train Data & $Acc_{Top1}$ & $Acc_{Top10}$ & $mAP@10$  \\ \hline
\multirow{8}{*}{Pix3D~\cite{sun2018pix3d}}        & ULIP~\cite{xue2023ulip}      & Point-BERT~\cite{yu2021pointbert}    & Shapenet~\cite{shapenet2015}       & 0.4/52.5     & 3.2/80.7      & 1.0/53.9  \\
                              &           &              & Ensembled~\cite{liu2023openshape}      & 0.9/59.1     & 4.3/85.1      & 1.7/60.5  \\
                              & OpenShape~\cite{liu2023openshape} & Point-BERT(S)~\cite{yu2021pointbert} & Shapenet~\cite{shapenet2015}       & 32.8/88.5    & 81.5/98.5     & 47.6/89.1 \\
                              &           & Point-BERT(L)~\cite{yu2021pointbert} & Ensembled~\cite{liu2023openshape}      & \cellcolor{green!20}\textbf{34.6/90.5 }   & \cellcolor{green!20}\textbf{83.3/97.8}     & \cellcolor{green!20}\textbf{50.5/90.5} \\
                              &           & SparseConv~\cite{choy2019minkowski}   & Shapenet~\cite{shapenet2015}       & 12.4/74.5    & 49.6/98.6     & 22.1/78.9 \\
                              &           &              & Ensembled~\cite{liu2023openshape}      & 16.4/89.9    & 61.3/99.3     & 28.5/88.5 \\
                              & ULIP2~\cite{ulip2}     & Point-BERT~\cite{yu2021pointbert}    & Ensembled~\cite{liu2023openshape}      & 0.7/57.3     & 4.2/82.1      & 1.6/60.4  \\
                              &           & PointNeXt~\cite{qian2022pointnext}    & Ensembled~\cite{liu2023openshape}      & 6.2/69.0     & 29.6/79.1     & 12.1/71.0 \\ \hline
\multirow{8}{*}{CompCars~\cite{yang2015compcars}}     & ULIP~\cite{xue2023ulip}      & Point-BERT~\cite{yu2021pointbert}     & Shapenet~\cite{shapenet2015}       & 0.7/-        & 10.1/-        & 2.6/-     \\
                              &           &              & Ensembled~\cite{liu2023openshape}      & 1.1/-        & 10.0/-        & 2.9/-     \\
                              & OpenShape~\cite{liu2023openshape} & Point-BERT(S)~\cite{yu2021pointbert}  & Shapenet~\cite{shapenet2015}       & 10.8/-       & \cellcolor{green!20}\textbf{47.7/-}        & \cellcolor{green!20}\textbf{20.1/-}    \\
                              &           & Point-BERT(L)~\cite{yu2021pointbert}  & Ensembled~\cite{liu2023openshape}      & \cellcolor{green!20}\textbf{10.9/-}       & 44.2/-        & 19.5/-    \\
                              &           & SparseConv~\cite{choy2019minkowski}   & Shapenet~\cite{shapenet2015}       & 3.8/-        & 22.6/-        & 8.1/-     \\
                              &           &              & Ensembled~\cite{liu2023openshape}      & 7.7/-        & 29.0/-        & 12.7/-    \\
                              & ULIP2~\cite{ulip2}     & Point-BERT~\cite{yu2021pointbert}    & Ensembled~\cite{liu2023openshape}      & 0.7/-        & 12.8/-        & 3.2/-     \\
                              &           & PointNeXt~\cite{qian2022pointnext}    & Ensembled~\cite{liu2023openshape}      & 0.5/-        & 11.3/-        & 3.1/-     \\ \hline
\multirow{8}{*}{StanfordCars~\cite{krause2013stanfordcars}} & ULIP~\cite{xue2023ulip}      & Point-BERT~\cite{yu2021pointbert}     & Shapenet~\cite{shapenet2015}       & 1.7/-        & 8.0/-         & 3.0/-     \\
                              &           &              & Ensembled~\cite{liu2023openshape}      & 1.8/-        & 8.6/-         & 3.3/-     \\
                              & OpenShape~\cite{liu2023openshape} & Point-BERT(S)~\cite{yu2021pointbert} ~\cite{yu2021pointbert} & Shapenet~\cite{shapenet2015}       & 11.2/-       & \cellcolor{green!20}\textbf{49.9/-}        & \cellcolor{green!20}\textbf{21.9/-}    \\
                              &           & Point-BERT(L)~\cite{yu2021pointbert}  & Ensembled~\cite{liu2023openshape}      & \cellcolor{green!20}\textbf{11.4/-}       & 49.1/-        & 21.8/-    \\
                              &           & SparseConv~\cite{choy2019minkowski}   & Shapenet~\cite{shapenet2015}       & 4.0/-        & 22.1/-        & 8.4/-     \\
                              &           &              & Ensembled~\cite{liu2023openshape}      & 5.6/-        & 33.1/-        & 12.2/-    \\
                              & ULIP2~\cite{ulip2}     & Point-BERT~\cite{yu2021pointbert}     & Ensembled~\cite{liu2023openshape}      & 1.7/-        & 8.4/-         & 3.4/-     \\
                              &           & PointNeXt~\cite{qian2022pointnext}    & Ensembled~\cite{liu2023openshape}      & 0.9/-        & 12.1/-        & 3.2/-     \\ \hline
\end{tabular}%
}
\caption[\textit{Zero-Shot} results for IBSR datasets]{Zero-Shot retrieval results for image-centered (aligned using multi-modal InfoNCE Eq.~\ref{eq:finetune_loss}) for IBSR datasets Pix3D\cite{sun2018pix3d}, CompCars\cite{yang2015compcars} and \cite{krause2013stanfordcars} \textit{(Notation: $\text{instance}/\text{class}$)}}
\label{tab:zeroshot_ibsr}
\end{table}
Evaluation on shape-centric datasets confirms that OpenShape~\cite{liu2023openshape} 
outperforms ULIP~\cite{xue2023ulip,ulip2}, with large-scale pre-training 
further improving zero-shot performance. The scaled Point-BERT~\cite{yu2021pointbert} 
achieves sufficient class-level accuracy for retrieval-based shape classification.
These trends persist on IBSR benchmarks (Table~\ref{tab:zeroshot_ibsr}), where 
Point-BERT~\cite{yu2021pointbert} encoders again perform best. However, zero-shot performance drops 
notably, particularly for instance-level retrieval. This gap stems from domain shift between synthetic pre-training data and real IBSR images. Pre-aligned models particularly struggle with instance-level retrieval compared to shape-centric datasets.

\begin{table}[htb!]
\centering
\resizebox{0.8\textwidth}{!}{%
\begin{tabular}{l|ll|ccc}
Dataset                     & Model        & Pre-train Data & $Acc_{Top1}$ & $Acc_{Top10}$ & $mAP@10$  \\ \hline
\multirow{4}{*}{Modelnet40~\cite{wu2015modelnet}} & Point-BERT(S)~\cite{yu2021pointbert}(ours) & Shapenet~\cite{shapenet2015}       & 45.1/90.2    & 88.4/98.7     & 59.3/88.3 \\
                            & Point-BERT(L)~\cite{yu2021pointbert}(ours) & Ensembled~\cite{liu2023openshape}      & \cellcolor{green!20}\textbf{57.3/92.2}    & \cellcolor{green!20}\textbf{92.5/99.1}     & \cellcolor{green!20}\textbf{69.1/88.9} \\
                            & SparseConv~\cite{choy2019minkowski}(ours)  & Shapenet~\cite{shapenet2015}       & 42.9/89.9    & 87.1/99.1     & 57.1/88.1 \\
                            &              & Ensembled~\cite{liu2023openshape}      & 45.4/91.1    & 88.6/99.1     & 59.3/88.8 \\ \hline
Pix3D~\cite{sun2018pix3d}                       & LFD~\cite{grabner2019location_fields}          & -              & 60.7/-       & 86.3/-        & -         \\
                            & HEG-TS~\cite{fu2020hard_example}      & -              & 74.9/-       & 95.0/-        & -         \\
                            & CMIC~\cite{yang2021ibsr}         & -              & 78.9/-       & 96.1/-        & -         \\
                            & SC-IBSR~\cite{song2024scibsr}      & -              & 80.2/-       & 96.9/-        & -         \\ \cline{2-6} 
                            & Point-BERT(S)~\cite{yu2021pointbert}(ours)& Shapenet~\cite{shapenet2015}       & 77.5/97.7     & 98.5/99.9        & 85.6/93.5    \\
                            & Point-BERT(L)~\cite{yu2021pointbert}(ours)& Ensembled~\cite{liu2023openshape}      & \cellcolor{green!20}\textbf{80.7/97.7}       & \cellcolor{green!20}\textbf{98.5/99.9}        & \cellcolor{green!20}\textbf{87.8/91.1}    \\
                            & SparseConv~\cite{choy2019minkowski}(ours)  & Shapenet~\cite{shapenet2015}       & 71.1/97.7     & 98.3/99.9     & 81.3/94.2    \\
                            &              & Ensembled~\cite{liu2023openshape}      & 69.6/97.4      & 97.8/99.8      & 80.0/94.1    \\ \hline
CompCars~\cite{yang2015compcars}                    & LFD~\cite{grabner2019location_fields}          & -              & 20.5/-       & 58.0/-        & -         \\
                            & HEG-TS~\cite{fu2020hard_example}       & -              & 67.1/-       & 93.7/-        & -         \\
                            & CMIC~\cite{yang2021ibsr}         & -              & 77.8/-       & 94.1/-        & -         \\
                            & SC-IBSR~\cite{song2024scibsr}      & -              & 78.7/-       & 94.2/-        & -         \\ \cline{2-6} 
                            & Point-BERT(S)~\cite{yu2021pointbert}(ours) & Shapenet~\cite{shapenet2015}       & 93.8/-       & 99.9/-        & 96.5/-    \\
                            & Point-BERT(L)~\cite{yu2021pointbert}(ours)& Ensembled~\cite{liu2023openshape}      & \cellcolor{green!20}\textbf{97.7/-}       & \cellcolor{green!20}\textbf{100.0/-}       & \cellcolor{green!20}\textbf{98.8/-}    \\
                            & SparseConv~\cite{choy2019minkowski}(ours)  & Shapenet~\cite{shapenet2015}       & 89.5/-       & 99.8/-        & 93.8/-    \\
                            &              & Ensembled~\cite{liu2023openshape}      & 85.1/-       & 99.2/-        & 90.5/-    \\ \hline
StanfordCars~\cite{krause2013stanfordcars}                & LFD~\cite{grabner2019location_fields}          & -              & 29.5/-       & 68.4/-        & -         \\
                            &HEG-TS~\cite{fu2020hard_example}       & -              & 68.4/-       & 92.1/-        & -         \\
                            & CMIC~\cite{yang2021ibsr}         & -              & 83.4/-       & 96.4/-        & -         \\
                            & SC-IBSR~\cite{song2024scibsr}      & -              & 84.3/-       & 97.1/-        & -         \\ \cline{2-6} 
                            & Point-BERT(S)~\cite{yu2021pointbert}(ours)& Shapenet~\cite{shapenet2015}       & 93.2/-       & 99.9/-        & 96.3/-    \\
                            & Point-BERT(L)~\cite{yu2021pointbert}(ours) & Ensembled~\cite{liu2023openshape}      & \cellcolor{green!20}\textbf{95.8/-}       & \cellcolor{green!20}\textbf{99.9/-}        & \cellcolor{green!20}\textbf{97.7/-}    \\
                            & SparseConv~\cite{choy2019minkowski}(ours)   & Shapenet~\cite{shapenet2015}       & 90.5/-       & 99.9/-        & 94.7/-    \\
                            &              & Ensembled~\cite{liu2023openshape}      & 88.4/-       & 99.9/-        & 93.6/-   
\end{tabular}%
}
\caption[Standard Retrieval results]{Retrieval results for OpenShape~\cite{liu2023openshape} models finetuned with multi-modal InfoNCE loss~\cite{oord2018infonce} on ModelNet~\cite{wu2015modelnet}, Pix3D~\cite{sun2018pix3d}, CompCars~\cite{yang2015compcars}, and StanfordCars~\cite{krause2013stanfordcars} \textit{(Notation: $\text{instance}/\text{class}$)}}
\label{tab:standard_performance}
\end{table}
\begin{table}[htb!]
\centering
\resizebox{0.9\textwidth}{!}{%
\begin{tabular}{l|l|cc|cc|cc}
\multirow{2}{*}{Dataset} & \multirow{2}{*}{Model} & \multicolumn{2}{c|}{$Acc_{Top1}$} & \multicolumn{2}{c|}{$Acc_{Top10}$} & \multicolumn{2}{c}{$mAP@10$} \\ \cline{3-8} 
                         &                        & Ref             & HCL(ours)             & Ref              & HCL(ours)            & Ref           & HCL(ours)         \\ \hline
Modelnet40~\cite{wu2015modelnet}               & Point-BERT(S)~\cite{yu2021pointbert}            & 24.7/73.2       & 24.4/72.2       & \cellcolor{green!20}\textbf{62.2/91.0}        & 61.0/91.4       & 35.8/74.5     & 35.0/73.7    \\
                         & Point-BERT(L)~\cite{yu2021pointbert}            & \cellcolor{green!20}\textbf{26.3/71.6}       &  25.6/71.0       & 61.5/90.2        & 60.1/89.7       & \cellcolor{green!20}\textbf{36.6/72.8}     & 35.8/71.6    \\
                         & SparseConv~\cite{choy2019minkowski}             & 24.5/71.2       & \cellcolor{green!20}\textbf{24.2/73.3 }      & 61.6/91.1        & \cellcolor{green!20}\textbf{61.2/91.4}       & 35.4/74.0     & \cellcolor{green!20}\textbf{35.2/74.5}    \\ \hline
Pix3D~\cite{sun2018pix3d}                    & Point-BERT(S)~\cite{yu2021pointbert}            & 27.4/84.6       & 27.8/85.9       & \cellcolor{green!20}\textbf{75.6/98.5}        & 73.9/98.5       & 43.4/84.6     & 41.8/84.8    \\
                         & Point-BERT(L)~\cite{yu2021pointbert}            & 30.1/84.9       & 29.4/81.8       & 79.5/87.7        & 73.2/98.6       & 46.0/84.9     & 43.2/81.6    \\
                         & SparseConv~\cite{choy2019minkowski}             & \cellcolor{green!20}\textbf{30.6/83.0 }      & \cellcolor{green!20}\textbf{33.5/87.3}       & 79.3/97.7        & \cellcolor{green!20}\textbf{97.0/99.2}       & \cellcolor{green!20}\textbf{45.7/83.4}     & \cellcolor{green!20}\textbf{47.8/86.4}    \\ \hline
CompCars~\cite{yang2015compcars}                 & Point-BERT(S)~\cite{yu2021pointbert}            & 10.6/-          & \cellcolor{green!20}\textbf{14.0/-}          & 41.5/-           & \cellcolor{green!20}\textbf{52.6/-}          & 18.3/-        & \cellcolor{green!20}\textbf{24.4/-}       \\
                         & Point-BERT(L)~\cite{yu2021pointbert}            & \cellcolor{green!20}\textbf{11.3/-}          & 11.2/-          & \cellcolor{green!20}\textbf{48.8/- }          & 44.0/-          & \cellcolor{green!20}\textbf{21.1/- }       & 19.3/-       \\
                         & SparseConv~\cite{choy2019minkowski}             & 9.2/-           & 9.0/-           & 39.1/-           & 43.1/-          & 17.3/-        & 17.6/-       \\ \hline
StanfordCars~\cite{krause2013stanfordcars}             & Point-BERT(S)~\cite{yu2021pointbert}            & 11.5/-          & 10.5/-          & 45.1/-           & 43.9/-          & 20.2/-        & 19.0/-       \\
                         & Point-BERT(L)~\cite{yu2021pointbert}            & \cellcolor{green!20}\textbf{12.2/-}          & \cellcolor{green!20}\textbf{10.8/-}          & \cellcolor{green!20}\textbf{45.9/-}           & 43.4/-          & \cellcolor{green!20}\textbf{21.2/-}        & \cellcolor{green!20}\textbf{19.6/- }      \\
                         & SparseConv~\cite{choy2019minkowski}             & 9.6/-           & 9.0/-           & 44.0/-           & \cellcolor{green!20}\textbf{44.4/- }         & 18.6/-        & 18.0/-      
\end{tabular}%
}
\caption[Zeroshot Retrieval with HCL training]{Zero-shot retrieval on ModelNet40~\cite{wu2015modelnet}, Pix3D~\cite{sun2018pix3d}, CompCars~\cite{yang2015compcars}, and StanfordCars~\cite{krause2013stanfordcars}, trained on ShapeNet~\cite{shapenet2015}. InfoNCE (Ref) vs. HCL (ours). \colorbox{green!20}{Highlighting} the best model for each loss objective. \textit{(Notation: $\text{instance}/\text{class}$)}} 
\label{tab:hcl_zeroshot}
\end{table}
\subsection{Standard Retrieval with InfoNCE loss}
\label{ssec:nce_standard}
Based on zero-shot results (\Cref{ssec:nce_zeroshot}), we fine-tune the four 
best-performing OpenShape~\cite{liu2023openshape} models on datasets with 
available train/test splits: all IBSR benchmarks and ModelNet40~\cite{wu2015modelnet}. 
Results in \Cref{tab:standard_performance} enable comparison with existing IBSR methods.
On ModelNet40, zero-shot performance trends persist after fine-tuning. 
Point-BERT(L)~\cite{yu2021pointbert} achieves the best instance-level 
$Acc_{\mathrm{Top}1}$ ($+12\%$ over other models), though this gap narrows 
to 4\% at $Acc_{\mathrm{Top}10}$. Class-level performance saturates at 
${\sim}90\%$ across all models.
On IBSR benchmarks, fine-tuned models show competitive or superior performance. 
For Pix3D~\cite{sun2018pix3d}, only scaled Point-BERT~\cite{yu2021pointbert} matches SOTA at 
$Acc_{\mathrm{Top}1}$, but all models surpass existing methods at 
$Acc_{\mathrm{Top}10}$. On CompCars~\cite{yang2015compcars} and 
StanfordCars~\cite{krause2013stanfordcars}, our models outperform prior 
work at $Acc_{\mathrm{Top}1}$ and nearly saturate at $Acc_{\mathrm{Top}10}$.
\subsection{Ablation Studies}
\begin{figure}[htb!]
\centering
\begin{subfigure}[t]{0.48\textwidth}
    \centering
    \pgfplotstableread{
model   A       B
0       34.0    45.1
1       30.6    57.3
2       31.6    45.4
}\mnaccone

\pgfplotstableread{
model   A       B
0       81.3    88.4
1       78.3    92.5
2       78.7    88.6
}\mnaccten

\pgfplotstableread{
model   A       B
0       48.7    59.3
1       45.1    69.1
2       46.0    59.3
}\mnmapten

\pgfplotstableread{
model   A       B
0       33.2    76.5
1       10.5    80.2
2       50.6    67.5
}\ptdaccone

\pgfplotstableread{
model   A       B
0       83.0    98.3
1       50.7    98.5
2       94.2    97.5
}\ptdaccten

\pgfplotstableread{
model   A       B
0       47.3    85.0
1       20.8    87.5
2       65.4    78.6
}\ptdmapten

\resizebox{\linewidth}{!}{%
\begin{tikzpicture}

\begin{groupplot}[group style={group size= 3 by 2},width=4cm]
    \nextgroupplot[
        title=$Acc_{Top1}$,
        ymin=0,
        ymax=100,
        ylabel=Modelnet40,
        xtick=data,
        xticklabels={PB(S), PB(L), SC},
        axis x line*=bottom,
        axis y line*=left,
        ymajorgrids,
        xticklabel style={yshift=-2pt},
        legend style={
            anchor=north east,
            at={(1,0)}
        },
        legend to name=bottom
    ]
        \addplot[
            only marks,
            mark=*,
            blue,
            mark size=2.5pt
        ] table [
            x=model,
            y=A
        ] {\mnaccone};

        \addplot[
            only marks,
            mark=square*,
            yellow,
            mark size=2.5pt
        ] table [
            x=model,
            y=B
        ] {\mnaccone};
    \nextgroupplot[
        title=$Acc_{Top10}$,
        ymin=0,
        ymax=100,
        xtick=data,
        xticklabels={PB(S), PB(L), SC},
        axis x line*=bottom,
        axis y line*=left,
        ymajorgrids,
        xticklabel style={yshift=-2pt},
    ]
        \addplot[
            only marks,
            mark=*,
            blue,
            mark size=2.5pt
        ] table [
            x=model,
            y=A
        ] {\mnaccten};

        \addplot[
            only marks,
            mark=square*,
            yellow,
            mark size=2.5pt
        ] table [
            x=model,
            y=B
        ] {\mnaccten};
    \nextgroupplot[
        title=$mAP@10$,
        ymin=0,
        ymax=100,
        xtick=data,
        xticklabels={PB(S), PB(L), SC},
        axis x line*=bottom,
        axis y line*=left,
        ymajorgrids,
        xticklabel style={yshift=-2pt},
    ]
        \addplot[
            only marks,
            mark=*,
            blue,
            mark size=2.5pt
        ] table [
            x=model,
            y=A
        ] {\mnmapten};

        \addplot[
            only marks,
            mark=square*,
            yellow,
            mark size=2.5pt
        ] table [
            x=model,
            y=B
        ] {\mnmapten};
    \nextgroupplot[
        ylabel=Pix3D,
        ymin=0,
        ymax=100,
        xtick=data,
        xticklabels={PB(S), PB(L), SC},
        axis x line*=bottom,
        axis y line*=left,
        ymajorgrids,
        xticklabel style={yshift=-2pt},
    ]
        \addplot[
            only marks,
            mark=*,
            blue,
            mark size=2.5pt
        ] table [
            x=model,
            y=A
        ] {\ptdaccone};

        \addplot[
            only marks,
            mark=square*,
            yellow,
            mark size=2.5pt
        ] table [
            x=model,
            y=B
        ] {\ptdaccone};
    \nextgroupplot[
        ymin=0,
        ymax=100,
        xtick=data,
        xticklabels={PB(S), PB(L), SC},
        axis x line*=bottom,
        axis y line*=left,
        ymajorgrids,
        xticklabel style={yshift=-2pt},
    ]
        \addplot[
            only marks,
            mark=*,
            blue,
            mark size=2.5pt
        ] table [
            x=model,
            y=A
        ] {\ptdaccten};

        \addplot[
            only marks,
            mark=square*,
            yellow,
            mark size=2.5pt
        ] table [
            x=model,
            y=B
        ] {\ptdaccten};
    \nextgroupplot[
        ymin=0,
        ymax=100,
        xtick=data,
        xticklabels={PB(S), PB(L), SC},
        axis x line*=bottom,
        axis y line*=left,
        ymajorgrids,
        xticklabel style={yshift=-2pt},
         legend style={
            anchor=north east,
            at={(1,-0.25)}
        },
        legend cell align={left},
        legend columns=2
    ]
        \addplot[
            only marks,
            mark=*,
            blue,
            mark size=2.5pt
        ] table [
            x=model,
            y=A
        ] {\ptdmapten};
        \addlegendentry{w/o pre-training}

        \addplot[
            only marks,
            mark=square*,
            yellow,
            mark size=2.5pt
        ] table [
            x=model,
            y=B
        ] {\ptdmapten};
        \addlegendentry{pre-trained}
\end{groupplot}

\end{tikzpicture}
}
    \caption{Instance retrieval on ModelNet40 \cite{wu2015modelnet} and Pix3D \cite{sun2018pix3d} comparing pre-trained vs. non-pre-trained OpenShape \cite{liu2023openshape} models: Point-BERT(S)~\cite{yu2021pointbert}, Point-BERT(L)~\cite{yu2021pointbert}, and SparseConv~\cite{choy2019minkowski} (PB(S), PB(L), SC).}
    \label{fig:rpt}
\end{subfigure}
\hfill
\begin{subfigure}[t]{0.48\textwidth}
    \centering
    \pgfplotstableread{
model   SR      SRA     HCL     HCLA
0       34.0    44.7    37.4    47.0
1       30.6    57.3    38.0    58.7
2       31.6    45.8    32.4    47.3
}\mnaccone

\pgfplotstableread{
model   SR       SRA    HCL     HCLA
0       81.3     88.3   81.6    88.2
1       78.3     93.0   83.5    92.3 
2       78.7     88.2   78.6    87.6
}\mnaccten

\pgfplotstableread{
model   SR      SRA     HCL     HCLA
0       48.7    59.1    51.3    60.7
1       45.1    69.7    53.8    70.4
2       46.0    59.5    46.5    60.5
}\mnmapten

\pgfplotstableread{
model   SR      SRA     HCL     HCLA
0       33.2    76.5    34.7    74.0
1       10.5    80.2    17.3    80.5
2       50.6    70.7    50.9    71.2
}\ptdaccone

\pgfplotstableread{
model   SR       SRA    HCL     HCLA
0       83.0    98.3    82.5    97.8
1       50.7    98.5    58.6    97.3
2       94.2    97.5    91.5    97.3
}\ptdaccten

\pgfplotstableread{
model   SR       SRA    HCL     HCLA
0       47.3    85.0    46.7    85.2
1       20.8    87.5    28.4    87.2
2       65.4    78.6    64.2    79.8
}\ptdmapten

\resizebox{\linewidth}{!}{%
\begin{tikzpicture}

\begin{groupplot}[group style={group size= 3 by 2},width=4cm]
    \nextgroupplot[
        title=$Acc_{Top1}$,
        ymin=20,
        ymax=70,
        ylabel=Modelnet40,
        xtick=data,
        xticklabels={PB(S), PB(L), SC},
        axis x line*=bottom,
        axis y line*=left,
        ymajorgrids,
        xticklabel style={yshift=-2pt},
        legend style={
            anchor=north east,
            at={(1,0)}
        },
        legend to name=bottom
    ]
        \addplot[
            only marks,
            mark=*,
            blue,
            mark size=2.5pt
        ] table [
            x=model,
            y=SR
        ] {\mnaccone};

        \addplot[
            only marks,
            mark=square*,
            yellow,
            mark size=2.5pt
        ] table [
            x=model,
            y=SRA
        ] {\mnaccone};

        \addplot[
            only marks,
            mark=triangle*,
            green,
            mark size=2.5pt
        ] table [
            x=model,
            y=HCL,
        ] {\mnaccone};

        \addplot[
            only marks,
            mark=pentagon*,
            magenta,
            mark size=2.5pt
        ] table [
            x=model,
            y=HCLA
        ] {\mnaccone};
    \nextgroupplot[
        title=$Acc_{Top10}$,
        ymin=75,
        ymax=100,
        xtick=data,
        xticklabels={PB(S), PB(L), SC},
        axis x line*=bottom,
        axis y line*=left,
        ymajorgrids,
        xticklabel style={yshift=-2pt},
    ]
        \addplot[
            only marks,
            mark=*,
            blue,
            mark size=2.5pt
        ] table [
            x=model,
            y=SR
        ] {\mnaccten};

        \addplot[
            only marks,
            mark=square*,
            yellow,
            mark size=2.5pt
        ] table [
            x=model,
            y=SRA
        ] {\mnaccten};

         \addplot[
            only marks,
            mark=triangle*,
            green,
            mark size=2.5pt
        ] table [
            x=model,
            y=HCL,
        ] {\mnaccten};

        \addplot[
            only marks,
            mark=pentagon*,
            magenta,
            mark size=2.5pt
        ] table [
            x=model,
            y=HCLA
        ] {\mnaccten};
    \nextgroupplot[
        title=$mAP@10$,
        ymin=40,
        ymax=75,
        xtick=data,
        xticklabels={PB(S), PB(L), SC},
        axis x line*=bottom,
        axis y line*=left,
        ymajorgrids,
        xticklabel style={yshift=-2pt},
    ]
        \addplot[
            only marks,
            mark=*,
            blue,
            mark size=2.5pt
        ] table [
            x=model,
            y=SR
        ] {\mnmapten};

        \addplot[
            only marks,
            mark=square*,
            yellow,
            mark size=2.5pt
        ] table [
            x=model,
            y=SRA
        ] {\mnmapten};

        \addplot[
            only marks,
            mark=triangle*,
            green,
            mark size=2.5pt
        ] table [
            x=model,
            y=HCL,
        ] {\mnmapten};

        \addplot[
            only marks,
            mark=pentagon*,
            magenta,
            mark size=2.5pt
        ] table [
            x=model,
            y=HCLA
        ] {\mnmapten};

    \nextgroupplot[
        ylabel=Pix3D,
        ymin=0,
        ymax=100,
        xtick=data,
        xticklabels={PB(S), PB(L), SC},
        axis x line*=bottom,
        axis y line*=left,
        ymajorgrids,
        xticklabel style={yshift=-2pt},
    ]
        \addplot[
            only marks,
            mark=*,
            blue,
            mark size=2.5pt
        ] table [
            x=model,
            y=SR
        ] {\ptdaccone};

        \addplot[
            only marks,
            mark=square*,
            yellow,
            mark size=2.5pt
        ] table [
            x=model,
            y=SRA
        ] {\ptdaccone};
        \addplot[
            only marks,
            mark=triangle*,
            green,
            mark size=2.5pt
        ] table [
            x=model,
            y=HCL,
        ] {\ptdaccone};

        \addplot[
            only marks,
            mark=pentagon*,
            magenta,
            mark size=2.5pt
        ] table [
            x=model,
            y=HCLA
        ] {\ptdaccone};
    \nextgroupplot[
        ymin=0,
        ymax=100,
        xtick=data,
        xticklabels={PB(S), PB(L), SC},
        axis x line*=bottom,
        axis y line*=left,
        ymajorgrids,
        xticklabel style={yshift=-2pt},
    ]
        \addplot[
            only marks,
            mark=*,
            blue,
            mark size=2.5pt
        ] table [
            x=model,
            y=SR
        ] {\ptdaccten};

        \addplot[
            only marks,
            mark=square*,
            yellow,
            mark size=2.5pt
        ] table [
            x=model,
            y=SRA
        ] {\ptdaccten};

        \addplot[
            only marks,
            mark=triangle*,
            green,
            mark size=2.5pt
        ] table [
            x=model,
            y=HCL,
        ] {\ptdaccten};

        \addplot[
            only marks,
            mark=pentagon*,
            magenta,
            mark size=2.5pt
        ] table [
            x=model,
            y=HCLA
        ] {\ptdaccten};
    \nextgroupplot[
        ymin=0,
        ymax=100,
        xtick=data,
        xticklabels={PB(S), PB(L), SC},
        axis x line*=bottom,
        axis y line*=left,
        ymajorgrids,
        xticklabel style={yshift=-2pt},
        legend style={
            anchor=north east,
            at={(1,-0.25)}
        },
        legend cell align={left},
        legend columns=2
    ]
        \addplot[
            only marks,
            mark=*,
            blue,
            mark size=2.5pt
        ] table [
            x=model,
            y=SR
        ] {\ptdmapten};
        \addlegendentry{NCE w/o pre-training}

        \addplot[
            only marks,
            mark=square*,
            yellow,
            mark size=2.5pt
        ] table [
            x=model,
            y=SRA
        ] {\ptdmapten};
        \addlegendentry{NCE pre-trained}

        \addplot[
            only marks,
            mark=triangle*,
            green,
            mark size=2.5pt
        ] table [
            x=model,
            y=HCL,
        ] {\ptdmapten};
        \addlegendentry{HCL(ours) w/o pre-training}

        \addplot[
            only marks,
            mark=pentagon*,
            magenta,
            mark size=2.5pt
        ] table [
            x=model,
            y=HCLA
        ] {\ptdmapten};
        \addlegendentry{HCL(ours) pre-trained}
\end{groupplot}

\end{tikzpicture}}
    \caption{Instance retrieval performance on ModelNet40~\cite{wu2015modelnet} and Pix3D~\cite{sun2018pix3d}. Comparison of InfoNCE~\cite{oord2018infonce} vs. HCL (ours) with/without pre-training. Models: PB(S), PB(L), SC (OpenShape~\cite{liu2023openshape}).}
    \label{fig:hcl_sr}
\end{subfigure}
\caption{Comparison of pre-training and loss function effects on instance retrieval.}
\label{fig:combined}
\end{figure}
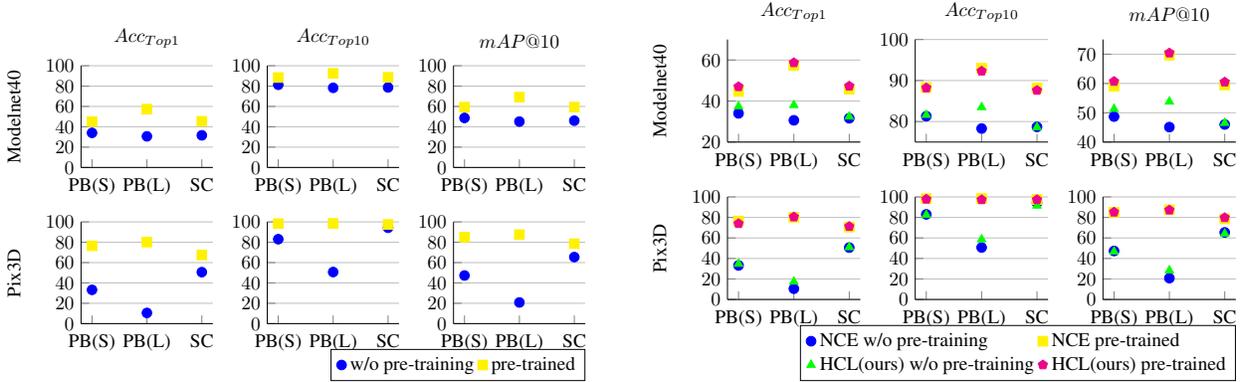

\noindent\textbf{Role of Pre-Training.}
\label{ssec:pre-training}
To quantify the impact of pre-alignment, we train Point-BERT(S)~\cite{yu2021pointbert}, Point-BERT(L)~\cite{yu2021pointbert} and SparseConv~\cite{choy2019minkowski} 
from scratch using identical settings as \Cref{ssec:nce_standard}.
As shown in Fig.~\ref{fig:rpt}, pre-aligned models consistently outperform their 
non-pre-aligned counterparts across both datasets and all metrics. The benefit 
is more pronounced on Pix3D~\cite{sun2018pix3d} than ModelNet40~\cite{wu2015modelnet}: 
Point-BERT(L)~\cite{yu2021pointbert} shows the largest gap (\textbf{80\% vs.\ 11\%} $Acc_{\mathrm{Top}1}$ on Pix3D~\cite{sun2018pix3d}), 
while SparseConv~\cite{choy2019minkowski} benefits least (17\% difference at $Acc_{\mathrm{Top}1}$, 
diminishing to 3\% at $Acc_{\mathrm{Top}10}$).
Notably, without pre-training, model performance converges—differences shrink 
to ${\sim}4\%$ at $Acc_{\mathrm{Top}1}$—whereas pre-alignment amplifies the 
gap to ${\sim}12\%$. At $Acc_{\mathrm{Top}10}$, non-pre-trained models approach 
pre-aligned performance, suggesting pre-training primarily improves fine-grained 
ranking.
\noindent\textbf{HCL effects on Zero-Shot performance.}
\label{ssec:hcl_zeroshot}
To evaluate the effect of hard negative learning on zero-shot performance (see Table~\ref{tab:hcl_zeroshot}), 
we pre-align point-cloud~\cite{yu2021pointbert, choy2019minkowski} encoders to image-encoders on ShapeNet~\cite{shapenet2015} using either standard 
InfoNCE loss \cite{oord2018infonce} (Ref) or our hard contrastive loss (HCL), keeping all other 
hyperparameters fixed. Models are evaluated on the same benchmarks as 
\Cref{ssec:nce_zeroshot}. Table~\ref{tab:hcl_zeroshot} shows that across ModelNet40~\cite{wu2015modelnet}, Pix3D~\cite{sun2018pix3d}, CompCars~\cite{yang2015compcars}, and StanfordCars~\cite{krause2013stanfordcars}, hard-negative learning yields scores comparable to the InfoNCE loss~\cite{oord2018infonce} for zero-shot retrieval.
\noindent\textbf{HCL Effects on Standard Retrieval.}
\label{ssec:hcl_standard}
HCL encourages instance-level discrimination by 
emphasizing hard negatives close to the anchor. We compare Point-BERT(S)~\cite{yu2021pointbert}, 
Point-BERT(L)~\cite{yu2021pointbert} and 
SparseConv~\cite{choy2019minkowski} trained with InfoNCE~\cite{oord2018infonce} vs. HCL, 
evaluating both pre-aligned (\Cref{ssec:nce_standard}) and non-pre-aligned 
(\Cref{ssec:pre-training}) settings.
As shown in Fig.~\ref{fig:hcl_sr}, HCL never degrades performance. On Pix3D~\cite{sun2018pix3d}, 
effects are negligible across all configurations. On ModelNet40~\cite{wu2015modelnet}, SparseConv~\cite{choy2019minkowski} 
shows no significant change, while Point-BERT~\cite{yu2021pointbert} models benefit notably. 
Without pre-alignment, Point-BERT(S)~\cite{yu2021pointbert} improves from 34.0\% to 37.4\% 
$Acc_{\mathrm{Top}1}$; Point-BERT(L)~\cite{yu2021pointbert} shows larger gains (\textbf{30.6\% → 38.0\%}). 
Pre-aligned models show modest improvements at $Acc_{\mathrm{Top}1}$ 
(e.g., 57.3\% → 58.7\% for Point-BERT(L)~\cite{yu2021pointbert}), with similar trends for $mAP@10$.
In summary, HCL provides consistent gains for Point-BERT~\cite{yu2021pointbert} architectures, 
particularly when training from scratch.
\subsection{Qualitative analysis}
\begin{figure}[htb!]
    \centering
    \begin{subfigure}{0.9\textwidth}
    \includegraphics[width=\textwidth]{media/p3d_os_pb_xl_zero_shot_3_neil.pdf}
    \caption{Zero-Shot}
    \end{subfigure}
    \begin{subfigure}{0.9\textwidth}
    \includegraphics[width=\textwidth]{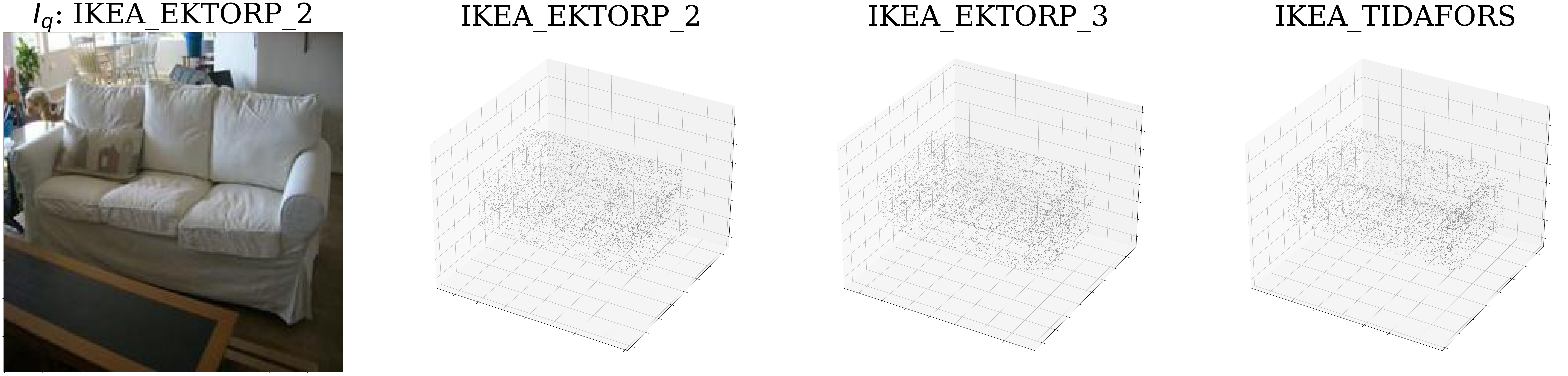}
    \caption{Standard Retrieval (InfoNCE)}
    \end{subfigure}
    \begin{subfigure}{0.9\textwidth}
    \includegraphics[width=\textwidth]{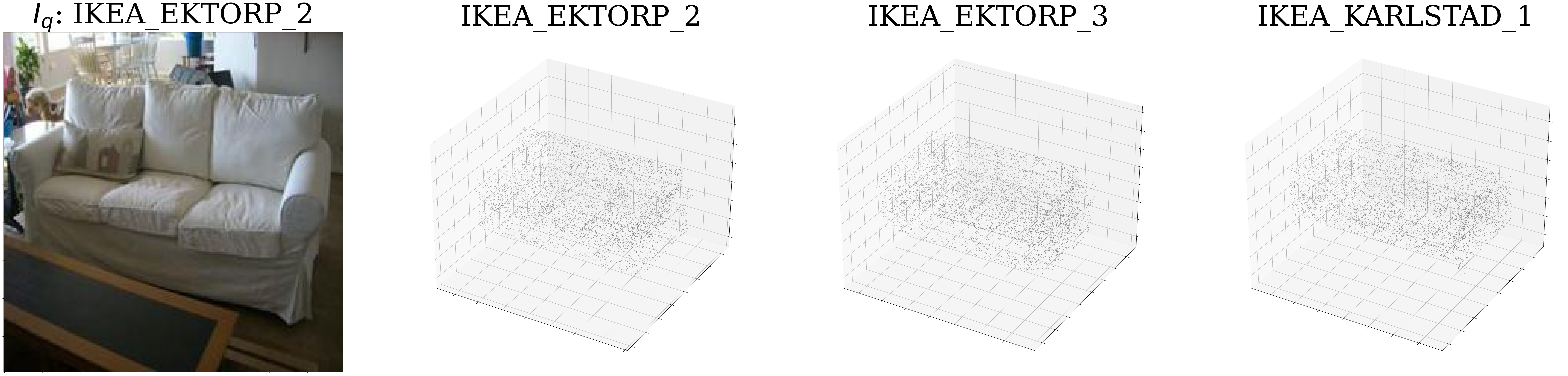}
    \caption{Standard Retrieval (HCL)}
    \end{subfigure}
    \caption{Qualitative results a given query image \textit{($I_g$)} of the IKEA\_EKTORP\_2 sofa from \cite{sun2018pix3d} comparing retrieved by the scaled Point-BERT~\cite{yu2021pointbert} model by OpenShape\cite{liu2023openshape} in its pre-aligned version (a), after fine-tuning on Pix3D \cite{sun2018pix3d} using the InfoNCE~\cite{oord2018infonce} loss (b) and after fine-tuning using our multi-modal HCL loss (c).}
    \label{fig:qualitative_analysis}
\end{figure}

Fig.~\ref{fig:qualitative_analysis} shows a representative IBSR example comparing zero-shot retrieval (with encoders aligned via InfoNCE loss~\cite{oord2018infonce}) and standard retrieval (fine-tuned with InfoNCE~\cite{oord2018infonce}/HCL). The standard retrieval achieves higher $Acc_{\mathrm{Top}1}$ accuracy and correctly retrieves IKEA\_EKTORP\_2 sofa from \cite{sun2018pix3d}, illustrating that fine-tuning improves $Acc_{\mathrm{Top}1}$ performance and yields clearer qualitative gains.

\section{Conclusion}
We presented a point cloud-based IBSR approach that preserves geometric detail while enabling effective transfer learning. Pre-aligned encoders support zero-shot retrieval and data-efficient fine-tuning. Our multi-modal HCL with hard negative sampling significantly improves instance-level discrimination, yielding substantial gains—particularly for Point-BERT~\cite{yu2021pointbert} in both fine-tuning and training from scratch.
Notably, our method reaches near-ceiling performance on several established benchmarks~\cite{wu2015modelnet, sun2018pix3d, yang2015compcars, krause2013stanfordcars}, particularly in terms of $Acc_{Top10}$. Rather than suggesting the problem of IBSR is entirely solved, these results highlight a maturation of performance on existing datasets. This underscores the critical need for more challenging, real-world benchmarks--such as OmniObject3D \cite{wu2023omniobject3d}--to continue pushing the boundaries of instance-level 3D discrimination.\\
\noindent\textbf{Future work.} Current benchmarks appear near saturation, motivating more challenging datasets. Future work includes multi-task pre-alignment (pose estimation, detection, segmentation) and domain-specific validation in robotics and augmented reality.

\bibliographystyle{splncs04}
\bibliography{main}

\end{document}